\documentclass[runningheads]{llncs}
\renewcommand{\figurename}{Figure}

\usepackage[utf8]{inputenc}
\usepackage[T1]{fontenc}
\usepackage{comment}
\usepackage{graphicx}
\usepackage{amsmath}
\usepackage{float}
\usepackage{multirow}
\usepackage{multicol}
\usepackage{marvosym}
\usepackage{ifsym}
\usepackage{amssymb}
\usepackage{textcomp}
\usepackage{orcidlink}
\usepackage{subcaption}
\usepackage{url}
\usepackage{algorithm}
\usepackage{algpseudocode}
\usepackage{longtable}
\usepackage{booktabs}
\usepackage{array}
\usepackage{ragged2e}
\usepackage{makecell}
\usepackage{tabularx}
\usepackage{hyperref}
\usepackage{graphicx}

\usepackage{float}
\usepackage{subcaption}
\makeatletter
\newenvironment{breakablealgorithm}
  {%
   \begin{center}
     \refstepcounter{algorithm}%
     \hrule height.8pt depth0pt \kern2pt%
     \renewcommand{\caption}[2][\relax]{%
       {\raggedright\textbf{\ALG@name~\thealgorithm} ##2\par}%
       \ifx\relax##1\relax
         \addcontentsline{loa}{algorithm}{\protect\numberline{\thealgorithm}##2}%
       \else
         \addcontentsline{loa}{algorithm}{\protect\numberline{\thealgorithm}##1}%
       \fi
       \kern2pt\hrule\kern2pt
     }
  }%
  {%
     \kern2pt\hrule\relax%
   \end{center}
  }
\makeatother

\usepackage[table]{xcolor}
\usepackage[autostyle=false, style=english]{csquotes}\MakeOuterQuote{"}
\renewcommand{\arraystretch}{1.25}

\begin{document}
\title{Reputation-driven Cooperation in Lattice-based Decentralized Federated Learning through Evolutionary Game Theory}
\titlerunning{Reputation-Driven Cooperation in Lattice-Based EGT-DFL}
%

\author{Phuc Hoang Truong Huynh\inst{1,2} \,\orcidlink{0009-0003-0731-3878} 
\and 
Dung Tran Vinh\inst{1,2} \,\orcidlink{0009-0005-9206-0944}
\and
Khoa Duc Anh Lam\inst{1,2} \,\orcidlink{0009-0002-3010-0822}
\and
An Nghiem Nguyen Truong\inst{1,2} \, \orcidlink{0009-0006-7194-2670}
\and 
Uyen Nha Tran Bui\inst{1,2} \, \orcidlink{}
\and
Khang Nguyen Dinh\inst{1,2} \, \orcidlink{https://orcid.org/0009-0004-1412-3607}
\and
Bao Nguyen Le Gia\inst{1,2} \,\orcidlink{0009-0008-9802-1307}
\and
Minh Le Nguyen Nhat\inst{1,2} \,\orcidlink{0009-0006-7622-4803}
\and
Manh Hong Duong\inst{3},\orcidlink{0000-0002-3095-7714}
\and
The Anh Han\inst{4*},\orcidlink{0000-0002-4361-0795}
\and
Thi Ai Thao Nguyen\inst{1,2*},\orcidlink{0000-0002-3111-0670}
\and
Le Hong Trang\inst{1,2}\thanks{Corresponding author}\,\orcidlink{0000-0002-6011-2261}
}
\authorrunning{Phuc Hoang et at.}

\institute{
Faculty of Computer Science and Engineering, Ho Chi Minh City University of Technology (HCMUT), Ho Chi Minh City, Vietnam\\
\email{\{hthphuc.sdh241, dung.tranvinh2005, khoa.lamgeo07, an.nghiemngtruong, uyen.buiangiang07, khang.nguyen2006siu, bao.nguyenbaobk4105, minh.lecomsci, thaonguyen, lhtrang\}@hcmut.edu.vn}
\and
Vietnam National University Ho Chi Minh City, Vietnam.
\and
School of Mathematics, University of Birmingham, Birmingham, United Kingdom
\and
School of Computing, Engineering and Digital Technologies, Teesside University, United Kingdom\\
\email{t.han@tees.ac.uk}
}
\maketitle              
\begin{abstract}
Decentralized Federated Learning (DFL) has emerged as an optimal privacy-preserving solution; however, it remains vulnerable to opportunistic behaviors due to the absence of a central coordinator. Although evolutionary game theory (EGT) serves as a powerful framework for analyzing such behaviors, existing studies often assume that agents possess perfect rationality and maintain static strategies. To address these limitations, this article proposes a novel EGT framework designed to analyze strategic evolution and improve overall system performance. The primary contributions of this work are threefold: First, we model peer-to-peer (P2P) interactions on a lattice network structure under the assumption of bounded rationality. Second, we formulate a comprehensive payoff matrix that incorporates training costs, communication overhead, and cooperative rewards, while tailoring a strategy update rule that captures the dynamics of spatial propagation. Third, we integrate a reputation-based reward-and-punishment mechanism to effectively deter free-riding behaviors. The simulation results demonstrate that the framework significantly outperforms the baseline. Specifically, it increases average accuracy from approximately $70\%$ to $82\%$, elevates cooperation frequency to approach $100\%$ (compared to below $5\%$ in the baseline), and drops accuracy variance from around $0.40$ to $0.002$, thereby accelerating uniform convergence and ensuring system stability.

\end{abstract}
\section{Introduction}\label{sec1}

In recent years, Machine Learning (ML) has gained widespread popularity due to its versatile applications across various domains \cite{lecun2015deep}. However, the rapid proliferation of data-driven models has raised severe concerns regarding data privacy and security \cite{ML_Concern}. To address these challenges, Federated Learning (FL) has emerged as an alternative paradigm \cite{fedavg}, enabling multiple devices or agents to collaboratively train a shared model while retaining their raw data on local client devices \cite{leaf}. Despite its significant advantages in data protection, traditional FL—specifically Centralized Federated Learning (CFL)—remains vulnerable to inherent limitations, particularly the risk of a single point of failure and potential information leakage at the central coordinating server \cite{DFL}.

To overcome these bottlenecks, Decentralized Federated Learning has surfaced as a promising solution \cite{first}. By completely eliminating the central server and allowing agents to autonomously adjust their strategic behavior over a Peer-to-Peer (P2P) network \cite{forth}, DFL enhances the overall security and reliability of the system \cite{did}. Nevertheless, this shift toward a fully decentralized architecture introduces complex behavioral challenges. DFL systems rely fundamentally on voluntary participation and mutual cooperation among agents. Within this environment, self-interested agents continuously modify their strategies to maximize individual utility, frequently giving rise to opportunistic behaviors such as "free-riding" \cite{free_rider} and malicious poisoning attacks \cite{byzantine} intended to disrupt model performance. Consequently, DFL is susceptible to model degradation, slow convergence rates, and, in severe cases, total network collapse \cite{byzantine_fault}.

To mitigate these behavioral and cooperative obstacles in DFL, the application of Evolutionary Game Theory (EGT) becomes indispensable \cite{incentive_mechanism}. Conventional game-theoretic approaches often rely on the unrealistic assumption of fully rational agents. In contrast, EGT offers a more realistic framework by modeling agents with bounded rationality whose strategies dynamically evolve over time based on accumulated payoffs \cite{weibull1997evolutionary,bashir2026strategic}. Crucially, a vast majority of existing EGT studies in FL assume well-mixed populations with uniform random interactions. In real-world DFL scenarios, however, agents interact exclusively with their direct topological neighbors \cite{gossip}. The relative scarcity of literature modeling interactions on specific spatial topographies, such as structured lattices \cite{square_lattice}, leaves a significant research gap regarding the mechanisms of spatial clustering for cooperation \cite{game_graph,nowak1992spatial}. Therefore, integrating EGT with a spatial lattice topology plays a pivotal role in analyzing behavioral dynamics and enhancing the sustainable performance of DFL systems.

This paper proposes an evolutionary game-theoretic framework to investigate the behavioral dynamics of agents (nodes) in DFL. The primary contributions of this study are threefold:

\vspace{-1mm}
\begin{itemize}
    \item Modeling agent interactions in DFL: we construct a lattice-based network topology grounded in EGT. To the best of our knowledge, this is one of the first studies to explicitly model P2P interactions among agents under the assumption of bounded rationality.
    
    \item Payoff function design and strategy update rules: we formulate a novel model that incorporates local training costs, communication overheads, and the mutual benefits derived from collaborative training. In addition, we refine the local strategy update rules to capture the imitation and dynamics of strategy propagation across neighboring nodes in the network.
    
    \item Reputation-based incentive mechanisms: To mitigate the free-riding problem inherent in DFL networks, we incorporate a reputation-based incentive scheme to foster mutual cooperation. Based on this reputation metric, the system dynamically rewards agents maintaining high reputation scores while penalizing those with low scores.
\end{itemize}

The rest of the paper is organized as follows. Section \ref{sec2} introduces related work. Section \ref{sec3} provides background knowledge. Section \ref{sec4} proposes the methodology, including architecture, modeling and workflow. Section \ref{sec5} evaluates its application and effectiveness by simulation. Finally, Section \ref{sec6} concludes the paper.

\section{Related Work}\label{sec2}

Federated Learning was initially introduced to address the escalating data privacy and security concerns in traditional machine learning. However, FL is highly susceptible to opportunistic behaviors, most notably the ``free-riding'' problem, where resource-constrained nodes skip local training but absorb updated models from others \cite{yang2023multi}. Early attempts to mitigate this issue primarily relied on traditional incentive mechanisms grounded in classical game theory, such as Stackelberg games or contract theory \cite{xu2025teg}. For instance, Xu \textit{et al.} \cite{xu2025teg} introduced a Tripartite Evolutionary Game to model interactions among a central server, organizers, and collaborators to address partial free-riding. Similarly, Yang \textit{et al.} \cite{yang2023multi} utilized a system dynamics-based multi-player evolutionary game to stabilize FL systems by regulating central server rewards and punishments. While effective, these early centralized approaches assumed perfect rationality and global observability, constantly facing single-point-of-failure risks and privacy bottlenecks inherent to the central coordinating server.

To eliminate the dependency on a central server, the field naturally evolved toward DFL, allowing nodes to aggregate models autonomously via P2P gossip protocols. In this new decentralized context, nodes operate with bounded rationality, dynamically adjusting their strategies based on observed payoffs rather than global information. To model this behavioral shift, researchers increasingly adopted EGT. For example, R{\"o}der \textit{et al.} \cite{roder2025driving} modeled DFL client interactions using the Iterated Prisoner's Dilemma, employing Moran sampling as an evolutionary incentive mechanism to systematically exclude non-cooperative clients from the training process. Although this transition to EGT effectively addressed bounded rationality, these strategies were primarily evaluated in well-mixed populations or simplistic topologies, often overlooking the physical and structural constraints of real-world edge devices.

Recognizing the limitations of well-mixed assumptions, recent studies have begun to emphasize the impact of structured populations and spatial topologies on cooperative behavior. In real-world DFL, edge devices are bound to specific network structures and interact exclusively with their direct topological neighbors. Duong \textit{et al.} \cite{duong2026beyond} demonstrated through agent-based simulations on square lattices that local, neighborhood-based interventions significantly outperform global, population-wide schemes in terms of both social welfare and cost-efficiency. Their work highlighted a critical evolutionary phenomenon: local interactions induce spatial clustering of cooperators, enabling cooperative behaviors to survive and propagate even when local payoffs inherently favor defection \cite{duong2026beyond}.

Despite the rich historical progression from centralized EGT models \cite{yang2023multi} to decentralized EGT \cite{roder2025driving} and spatial cooperation \cite{duong2026beyond}, a comprehensive framework that explicitly models fully decentralized P2P interactions on a spatial lattice remains underexplored. Existing DFL incentive models lack the integration of localized imitation dynamics (e.g., the Fermi update rule) with practical FL expenditures, such as communication overhead and local training costs. To bridge this historical gap, this paper proposes a lattice-based Evolutionary-Game-Theoretic DFL architecture. By integrating a reputation-based incentive mechanism and neighborhood gossip averaging, our framework fosters the spatial clustering of cooperators, effectively mitigating the free-riding problem without relying on any central authority.

\section{Background }\label{sec3}
\subsection{Decentralized Federated Learning and the Free-Riding Problem}

Unlike traditional server-centric FL, DFL eliminates the single point of failure and communication bottlenecks by enabling edge nodes to collaborate via P2P network topologies \cite{DFL_foundation}. In a DFL architecture, global model consensus is achieved purely through local computation and iterative neighborhood interactions governed by Gossip Averaging protocols \cite{gossip}.

Formally, within our evolutionary game framework over a graph $\mathcal{G} = (\mathcal{V}, \mathcal{E})$, each node $i \in \mathcal{V}$ holds a local model weight vector $w_i^{(k)}$ at iteration round $k$. The primary objective of average consensus in DFL is for every node's model parameters to converge toward the exact global arithmetic mean:
\begin{equation}
    \bar{w} = \frac{1}{n} \sum_{i=1}^{n} w_i^{(0)}
\end{equation}

In a \textbf{synchronous neighborhood averaging} scheme over a regular network topology - specifically, a periodic square lattice where each node $i$ is connected to $|\mathcal{N}_i| = 4$ direct neighbors—all nodes concurrently mix their local model weights with their open neighborhood. The state transition for an individual node $i$ at iteration step $k+1$ is locally computed as:
\begin{equation}
    w_i^{(k+1)} = \frac{1}{5} \left( w_i^{(k)} + \sum_{j \in \mathcal{N}_i} w_j^{(k)} \right).
\end{equation}

Compactly, across the entire network, the global state transition is governed by a doubly-stochastic mixing matrix $W^{(k)} \in \mathbb{R}^{n \times n} $ \cite{gossip}:
\begin{equation}
    w^{(k+1)} = W^{(k)} w^{(k)}, 
\end{equation}
where the entries of $W^{(k)}$ for this 4-neighbor square lattice topology are defined as:
\begin{equation}
    W_{ij}^{(k)} = \begin{cases} 
    \frac{1}{5}, & \text{if } j = i \text{ or } j \in \mathcal{N}_i \\ 
    0, & \text{otherwise} 
    \end{cases}
\end{equation}
This matrix construction naturally satisfies $W_{ij}^{(k)} \ge 0$, $W_{ij}^{(k)} = 0$ for $(i,j) \notin \mathcal{E}$, $W^{(k)} \mathbf{1} = \mathbf{1}$, and $\mathbf{1}^T W^{(k)} = \mathbf{1}^T$. This guarantees that as iterations progress, the matrix product converges to $\lim_{k \to \infty} W^{(k-1)} \dots W^{(0)} = J$ (where $J = \frac{1}{n} \mathbf{1} \mathbf{1}^T$), driving $w_i^{(k)} \to \bar{w}$ for all nodes $i$ \cite{gossip}.

However, the P2P nature and non-excludable neighborhood mixing step inherently expose DFL to the \textit{Free-riding problem} \cite{free_rider}. Free-riders are self-interested or resource-constrained nodes that aim to reap the benefits of a well-trained global model without contributing their fair share of computational resources.

To contribute to global performance, a cooperative node ($S_i = C$) executes local model training (e.g., stochastic gradient descent) prior to gossip aggregation, expending computational and energy cost $c > 0$. In contrast, a defecting node ($S_i = D$) skips local training ($c = 0$) or sends stale model weights, yet still receives and averages the updated state vectors $w_j^{(k)}$ from its neighbors via $W^{(k)}$.
Formally, the net fitness payoff $P_i$ of node $i$ is modeled as a function of the global model performance benefit $B(\cdot)$ derived from neighborhood aggregation minus its incurred operational cost $c_i$:
\begin{equation}
    P_i = B\left(\{w_j^{(k)}\}_{j \in \mathcal{N}_i \cup \{i\}}\right) - c_i
\end{equation}
where $c_i = c$ if $S_i = C$, and $c_i = 0$ if $S_i = D$. Because $B(\cdot)$ is shared locally due to the non-excludable nature of gossip protocols, a defecting node avoids the cost $c$ while still exploiting the accuracy gains provided by its cooperative neighbors. Consequently, for equivalent neighborhood inputs, a free-riding node achieves a strictly higher net fitness payoff than its cooperative counterparts:
\begin{equation}
    P_i(D) > P_i(C)
\end{equation}

\subsection{Evolutionary Game Theory (EGT) on Network Topologies}
\noindent{\bf Game Formulation and Payoff Structure}.
Game theory provides the analytical framework to understand and predict strategic decision-making in interactive scenarios marked by conflicting priorities among players \cite{game_graph,weibull1997evolutionary}. For EGT, unlike classical game theory, which assumes fully rational players with complete information, EGT models boundedly rational agents who iteratively adjust their strategies based on observed payoffs over time \cite{hofbauer1998evolutionary}.
We model the pairwise interaction between edge devices in DFL as a symmetric two-player normal-form game $G = (A, A^T)$ \cite{game_graph}. Each node $i$ adopts a strategy $S_i(t) \in \{C, D\}$ at round $t$, where $C$ represents Cooperation (executing local Stochastic Gradient Descent (SGD) training) and $D$ represents Defection (free-riding or transmitting uninformative updates).
The expected payoff matrix $A$ governing the interaction between node $i$ and neighbor $j$ under strategy profile $(S_i, S_j)$ is formulated as:
\begin{equation}
    A = \begin{pmatrix}
    U_{CC} & U_{CD} \\
    U_{DC} & U_{DD}
    \end{pmatrix} = 
    \begin{pmatrix}
    Q(2) - c & Q(1) - c \\
    Q(1) & 0
    \end{pmatrix}
    \label{eq:payoff_matrix}
\end{equation}
where $Q(k)$ denotes the aggregated model performance gain derived from $k$ actively contributing neighbors ($k \in \{0, 1, 2\}$), satisfying $Q(2) > Q(1) > Q(0) = 0$. The parameter $c > 0$ represents the local resource expenditure per training round (e.g., energy, compute power, and bandwidth).

\noindent{\bf Topological Network Interactions}.
In traditional well-mixed populations, players interact uniformly at random. However, in DFL, edge devices are bound to structured networked topologies (e.g., regular lattices, scale-free, or small-world networks), interacting exclusively with their direct topological neighbors. Let $\mathcal{N}_i$ denote the set of all neighbors of node $i$. The cumulative utility $P_i(t)$ obtained by node $i$ at round $t$ is the sum of pairwise payoffs with its neighbors:
\begin{equation}
    P_i(t) = \sum_{j \in \mathcal{N}_i} u(S_i(t), S_j(t))
    \label{eq:cumulative_payoff},
\end{equation}
where $u(S_i(t), S_j(t))$ is the pairwise payoff derived from matrix $A$. This local interaction induces spatial clustering of cooperators, enabling mutual cooperation to survive against free-riders even when the local payoff favors defection ($U_{DC} > U_{CC}$) \cite{game_graph}

\subsection{Strategy Adoption and Fermi Imitation Rule}
To model how edge nodes adapt their behavioral strategies over consecutive training rounds, we incorporate the pairwise imitation dynamics from spatial evolutionary game theory \cite{square_lattice}. Under bounded rationality, a node $i$ does not possess global information; instead, at the end of round $t$, it randomly selects a direct neighbor $j \in \mathcal{N}_i$ as a role model, to evaluate performance and behavioral update. Node $i$ adopts neighbor $j$'s strategy $S_j(t)$ for the subsequent round $t+1$ according to the Fermi update rule borrowed from statistical physics \cite{square_lattice}:
\begin{equation}
    W(S_i \leftarrow S_j) = \frac{1}{1 + e^{\frac{P_i(t) - P_j(t)}{K}}} 
    \label{eq:fermi_rule}
\end{equation}
where $P_i(t)$ and $P_j(t)$ denote the cumulative payoffs acquired by nodes $i$ and $j$, respectively. The parameter $K > 0$ represents the selection intensity, quantifying the level of environmental noise or decision-making irrationality:
\begin{itemize}
    \item As $K \to 0^+$, strategy adaptation becomes purely deterministic, where node $i$ strictly imitates neighbor $j$ if and only if $P_j(t) > P_i(t)$.
    \item As $K \to \infty$, decision-making degrades into a random coin toss ($W \approx 0.5$), ignoring payoff differentials due to severe noise or perception errors.
\end{itemize}
By regulating $K$, the Fermi dynamics effectively capture how cooperative behavior (e.g., executing local SGD in federated learning) forms spatial clusters and survives against free-riders under varying environmental noise levels.

\section{Methodology}\label{sec4}
\subsection{Lattice-based Evolutionary-Game-Theoretic Decentralized Federated Learning}\label{architecture}

We construct a DFL system over a two-dimensional lattice network, which is composed of overlapping clusters consisting of one central node, referred to as the \emph{focal node}, and its four immediate neighbors. These clusters are regarded as overlapping because each node simultaneously acts as the focal node of its own local neighborhood and as a neighbor of up to four other focal nodes. Each node maintains its own model and possesses local data as well as heterogeneous computational capabilities. The nodes collaboratively participate in training, exchange model parameters, and merge model weights with their neighbors through a peer-to-peer mechanism. Therefore, the system removes the dependency on a single centralized aggregator~\cite{lian2017dpsgd,sun2021dfedavgm,hu2019segmentedgossip}. The exchange and merging processes propagate across clusters throughout the network and are repeated over multiple rounds until model convergence is achieved.

During this process, under the influence of EGT, each node adopts an individual strategy \(s \in \{C,D\}\), where the node decides whether to share (\emph{Cooperator}: \(C\)) or not to share (\emph{Defector}: \(D\)) its model parameters with neighboring nodes. This decision is made based on the benefit obtained and the cost incurred when a specific strategy is adopted. Since the conditions of individual nodes are heterogeneous, their strategic decisions may differ, which makes the problem more realistic and less idealized than the conventional formulation of DFL.

All nodes participating in the DFL system seek to obtain a sufficiently good model for solving their own local learning tasks. Therefore, they are willing to perform training whenever the received model weights are improved. However, whether they contribute their own effort to the collective community depends on their strategic behavior in the game. A payoff matrix is established to represent this trade-off and to provide a basis for explaining the strategic choices of each node.

\begin{figure}[H]
    \centering    \includegraphics[width=0.6\textheight]{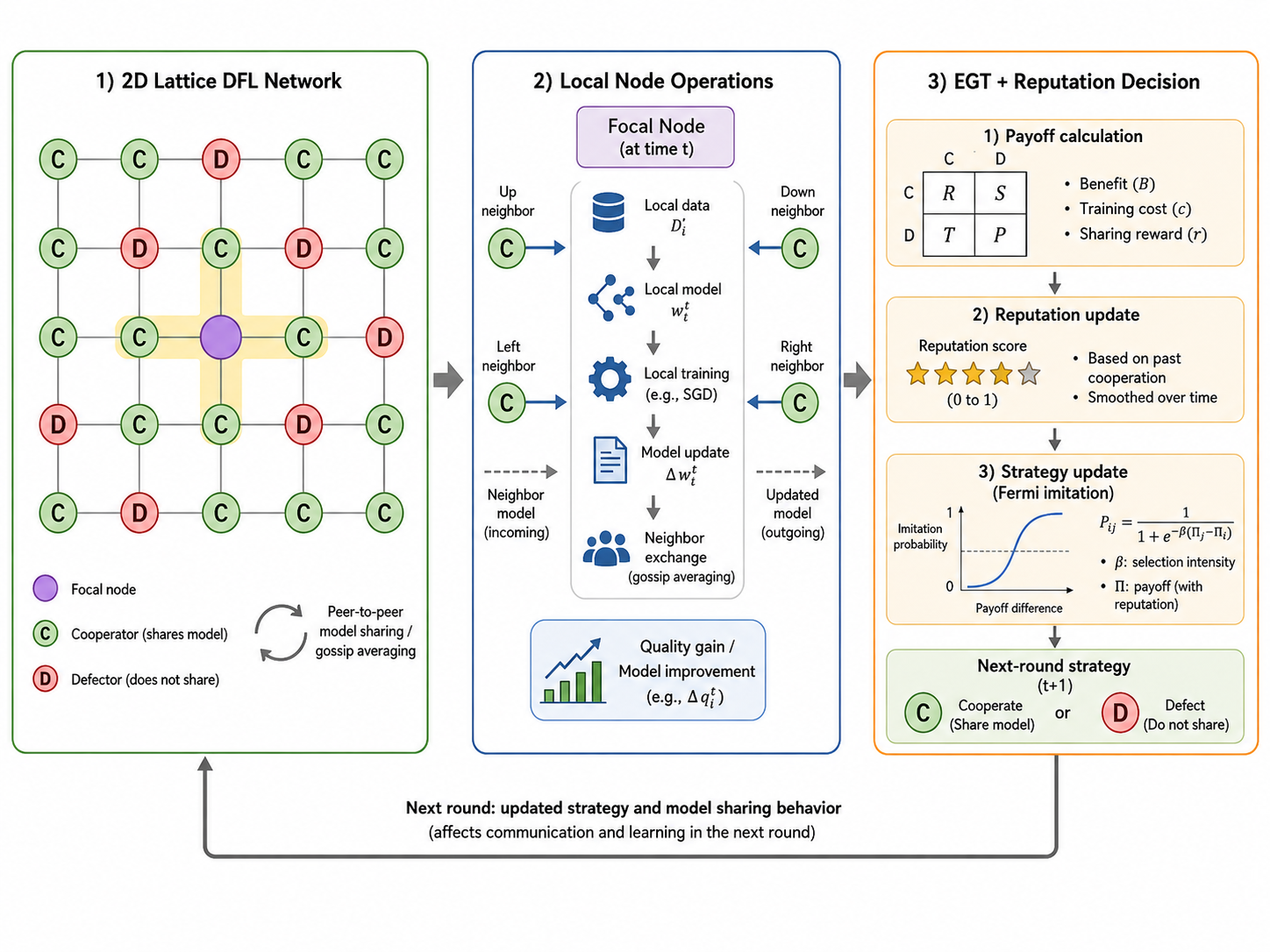}
    \vspace{-5mm}
    \caption{Overall architecture of the proposed lattice-based decentralized federated learning system, where nodes perform local training, peer-to-peer gossip averaging, reputation-aware payoff evaluation, and evolutionary strategy updating over a two-dimensional lattice network.}
    \label{fig:architecture}
\end{figure}

\subsection{Evolutionary-Game-Theoretic Modeling}

\begin{figure}[!ht]
    \centering    \includegraphics[width=0.6\textheight]{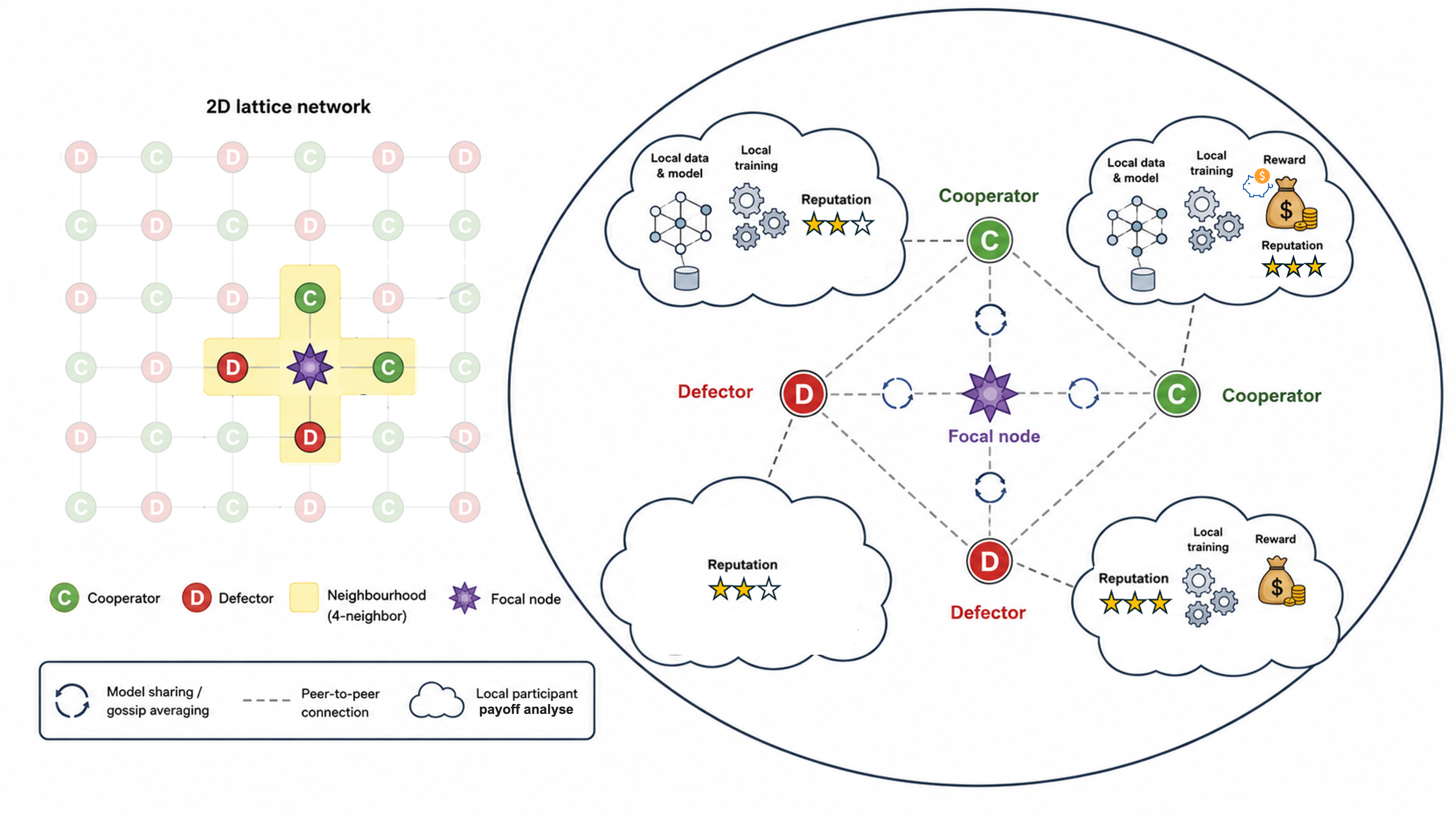}
    \caption{Payoff-based strategy assessment of a focal node when interacting with a neighboring node. The focal node compares its accumulated payoff with that of the neighbor, evaluates the potential benefits and costs associated with cooperation or defection, and probabilistically updates its strategy according to the relative payoff difference.}
    \label{fig:payoff}
\end{figure}

When participating in training, a focal node may face four interaction scenarios with respect to a neighboring node.

\begin{enumerate}
    \item \textbf{The focal node is a Cooperator and the neighbor is also a Cooperator (CC).}
    This represents the most desirable bilateral interaction, in which both nodes share the model weights obtained from their own training processes. This interaction provides a substantial improvement in the quality of the shared model, represented by \(+\Delta \mathrm{Q}\). Although both nodes incur communication overhead \((-O)\) and training cost \((-T)\) when retraining the model after receiving improved parameters, both of them are rewarded with positive reputation gains \((R^+)\).
    
    \newpage

    \item \textbf{The focal node is a Cooperator whereas the neighbor is a Defector (CD).}
    The Cooperator is a benevolent participant that is willing to share for the benefit of the community, thereby incurring the communication overhead \((-O)\). However, such cooperation may not always be reciprocated by the neighbor. In some cases, the reward does not come from the expected interacting node, but rather from community-level recognition of cooperative behavior, represented by reputation benefit \((R^+)\).

    \item \textbf{The focal node is a Defector whereas the neighbor is a Cooperator (DC).}
    In this case, the focal node behaves selfishly by receiving benefits without contributing back. The objective of maximizing individual utility prevents the focal node from sharing its own model parameters with the neighbor that has helped improve its model quality. This corresponds to the free-riding problem. The focal node trains for its own benefit and only returns the same parameters that it previously received. It accepts the associated costs \((-T,-O)\) and the reputation penalty \((R^-)\) because the benefit \((+\Delta \mathrm{Q}\)) obtained from free-riding may still be attractive.

    \item \textbf{The focal node is a Defector and the neighbor is also a Defector (DD).}
    In this case, both nodes refuse to share with each other and accept reputation degradation \((R^-)\) because they assume that the bilateral interaction does not provide sufficient benefit.
\end{enumerate}

\newpage
\noindent\textbf{Payoff Matrix}.
EGT adapts game theory to evolving populations, where strategies are dynamically adopted based on each participant's relative fitness, or payoff~\cite{nowak1992spatial,szabo2007evolutionary}.
Accordingly, a \(2 \times 2\) payoff matrix is constructed based on the two strategies \(s \in \{C,D\}\), representing the strategic trade-off between sharing and not sharing model parameters for improving the collective model. In this behavioral analysis, each payoff value quantifies the effectiveness of a bilateral interaction through carefully computed numerical values, which support each node in selecting an appropriate strategy in each training round.
The payoff rules governing two interacting nodes are summarized in Table~\ref{tab:baseline_payoff_matrix}.
\begin{table}[H]
\centering
\caption{Evolutionary-Game-Theoretic Assessment of the Baseline Payoff Matrix}
\label{tab:baseline_payoff_matrix}
\renewcommand{\arraystretch}{1.8}


\begin{tabular}{|c|c|c|}
\hline
\textbf{Strategy} & \textbf{$\mathbf{C_j}$} & \textbf{$\mathbf{D_j}$} \\
\hline
\textbf{$\mathbf{C_i}$} & $\alpha \left[\Delta Q_{C_iC_j}\right]^{+} - T_{C_i} - O_{C_i}$ & $- O_{C_i}$ \\
\hline
\textbf{$\mathbf{D_i}$} & $\beta \left[\Delta Q_{C_i}\right]^{+} - T_{D_i} - O_{D_i}$ & 0 \\
\hline
\end{tabular}
\end{table}

{\noindent\scriptsize\textit{Note}: Although $D_i$ doesn't share its model weights, it still propagates the unmodified weights received from $C_j$ (incurring $-O_{D_i}$) and performs local updates (incurring $-T_{D_i}$) to improve its model.
\par}

\vspace{2mm}
By observing the payoff matrix in Table.~\ref{tab:baseline_payoff_matrix}, the most attractive factor that encourages both the focal node and its neighbor to form a genuinely cooperative interaction is the improvement in the quality of the shared model, \(+\Delta Q\), which is amplified by the synergy coefficient \(\alpha>1\). This benefit is assigned to participants that make positive contributions according to their model-sharing ratio and their marginal contribution to the quality of the collective model. In contrast, the coefficient \(0\leq\beta\leq1\) characterizes the free-riding effect. Specifically, \(\beta\) represents the fraction of the model-quality benefit that a defecting focal node can retain by receiving the model parameters shared by a cooperative neighbor without reciprocally sharing its own updated parameters. A larger value of \(\beta\) indicates that the defector can appropriate a greater proportion of the neighbor's contribution, whereas a smaller value indicates that the exploitable benefit is limited. Accordingly, \(\beta=1\) represents complete appropriation of the received model-quality gain, while \(\beta=0\) indicates that
free-riding provides no effective model-quality benefit. Unlike
\(\alpha\), which captures the complementary value generated through
mutual cooperation, \(\beta\) does not represent a productive
contribution; rather, it quantifies the retained benefit obtained from one-sided cooperation.

Let \(i\) denote a focal node and \(j\in\mathcal{N}_i\) denote one of
its neighboring nodes. Let \(Q(w)\) denote the quality of model \(w\) (e.g., measured by its
accuracy on a shared validation set or by the negative value of its loss).
The marginal contribution of node \(i\) at round \(t\) can be measured
using the \emph{leave-one-out} method as follows:
\[
    \Delta Q_i^{(t)}
    = Q\!\left(w_{\mathcal{C}_t}^{(t)}\right)
    - Q\!\left(w_{\mathcal{C}_t \setminus \{i\}}^{(t)}\right),
\]
where \(\mathcal{C}_t\) denotes the set of contributing nodes at round
\(t\), \(w_{\mathcal{C}_t}^{(t)}\) is the model aggregated from all nodes
in \(\mathcal{C}_t\), and
\(w_{\mathcal{C}_t \setminus \{i\}}^{(t)}\) is the model aggregated after
excluding the contribution of node \(i\).
To prevent model updates that degrade model quality from receiving a
reward, the effective quality contribution of node \(i\) is defined as
\[
    q_i^{(t)}
    = \left[\Delta Q_i^{(t)}\right]^{+}
    = \max\!\left\{0,\Delta Q_i^{(t)}\right\}.
\]
At generation \(t\), each node adopts a strategy
\(s_i^{(t)}\in\{C,D\}\), where \(C\) and \(D\) denote cooperation and
defection, respectively. Let
\(
[x]^+\triangleq\max\{0,x\}
\)
denote the positive-part operator.
The pairwise payoff received by focal node \(i\)
from its interaction with neighbor \(j\) is defined as
\begin{equation}
\pi_{ij}^{(t)}
\left(s_i^{(t)},s_j^{(t)}\right)
=
\begin{cases}
\displaystyle
\alpha
\left[
    \Delta Q_{C_iC_j}^{(t)}
\right]^+
-
T_{C_i}^{(t)}
-
O_{C_i}^{(t)},
&
\left(s_i^{(t)},s_j^{(t)}\right)=(C,C),
\\[3mm]
\displaystyle
-
O_{C_i}^{(t)},
&
\left(s_i^{(t)},s_j^{(t)}\right)=(C,D),
\\[3mm]
\displaystyle
\beta
\left[
    \Delta Q_{C_i}^{(t)}
\right]^+
-
T_{D_i}^{(t)}
-
O_{D_i}^{(t)},
&
\left(s_i^{(t)},s_j^{(t)}\right)=(D,C),
\\[3mm]
0,
&
\left(s_i^{(t)},s_j^{(t)}\right)=(D,D).
\end{cases}
\label{eq:base-pairwise-payoff}
\end{equation}
Here, \(\Delta Q_{C_iC_j}^{(t)}\) denotes the model-quality improvement
resulting from mutual cooperation between nodes \(i\) and \(j\), while
\(\Delta Q_{C_i}^{(t)}\) denotes the model-quality benefit obtained by
the focal node in a \(DC\) interaction. The coefficients \(\alpha\) and
\(\beta\) represent the cooperative synergy and free-riding effects,
respectively. Moreover, \(T_i^{(t)}\) and \(O_i^{(t)}\) denote the
training and model-exchange costs.
The accumulated payoff of focal node \(i\) is
\begin{equation}
\Pi_i^{(t)}
=
\sum_{j\in\mathcal{N}_i}
\pi_{ij}^{(t)}
\left(
    s_i^{(t)},s_j^{(t)}
\right),
\label{eq:base-accumulated-payoff}
\end{equation}
and its evolutionary fitness is defined as
\begin{equation}
f_i^{(t)}
\triangleq
\Pi_i^{(t)}.
\label{eq:base-fitness}
\end{equation}

\noindent\textbf{Reputation}.
In addition to the quality of the shared model \(+\Delta \mathrm{Q}\), reputation is another important mechanism for guiding node behavior toward cooperative sharing that benefits the DFL system. The utility constrained by reputation is determined according to the reputation score accumulated over training rounds. A higher accumulated reputation score leads to a larger additional incentive, whereas a lower score reduces the benefit obtained by the node~\cite{kang2019reputation,zhan2022survey,zeng2021survey}.

Furthermore, the integration of reputation into the payoff framework serves another purpose. In the DFL context, the stopping condition of the system is not that the entire network reaches a homogeneous strategic state, but rather that the model reaches convergence. Therefore, even when the entire network reaches the best cooperative state \(C\), or falls into the worst conservative state \(D\), the DFL process must continue if the shared model has not yet achieved an acceptable level of consensus quality.

The design of the reputation score provides a mechanism for shifting the balance in favor of a node surrounded by an absorbing state, such as an all-defector neighborhood, or for imposing an unavoidable penalty that discourages a node from remaining trapped in a conservative strategy when its utility is decreasing. Reviving a node through a sharing-oriented strategy can generate a resonant propagation effect, thereby creating favorable conditions for the model quality to continue improving gradually toward convergence.

\vspace{-5mm}
\begin{table}[H]
\centering
\caption{Evolutionary-Game-Theoretic Assessment of the Reputation-based Payoff Matrix}
\label{tab:reputation_payoff_matrix}
\renewcommand{\arraystretch}{1.8}


\begin{tabular}{|c|c|c|}
\hline
\textbf{Strategy} & \textbf{$\mathbf{C_j}$} & \textbf{$\mathbf{D_j}$} \\
\hline
\textbf{$\mathbf{C_i}$} & $R^+_{C_i} (\alpha \left[\Delta Q_{C_iC_j}\right]^{+} - T_{C_i} - O_{C_i} )$ & $R^+_{C_i} (- O_{C_i})$ \\
\hline
\textbf{$\mathbf{D_i}$} & $R^-_{D_i} (\beta \left[\Delta Q_{C_i}\right]^{+} - T_{D_i} - O_{D_i})$ & $R^-_{D_i}$ \\
\hline
\end{tabular}
\end{table}

\vspace{-5mm}
The reputation value and its round-by-round update are defined as
\[
\begin{gathered}
R(x_{ij})=
\begin{cases}
    x_{ij}\,\delta_n, & x \geq 0, \\[3pt]
    \dfrac{x_{ij}}{\delta_n}, & x < 0,
\end{cases}
\\[8pt]
\delta_0=1, \qquad
\delta_n=
\begin{cases}
    \min\!\left\{\delta_{n-1}+r,\,\delta_{\max}\right\},
        & \text{if } s_{ij}=C, \\[3pt]
    \max\!\left\{\delta_{n-1}-r,\,\delta_{\min}\right\},
        & \text{if } s_{ij}=D.
\end{cases}
\end{gathered}
\]
When reputation is incorporated (Table. \ref{tab:reputation_payoff_matrix}), let
\(R_{C_i}^{+,(t)}\) denote the positive reputation modifier associated
with cooperative behavior, and let \(R_{D_i}^{-,(t)}\) denote the
reputation modifier associated with defective behavior. The
reputation-adjusted pairwise payoff is defined as
\begin{equation}
\widetilde{\pi}_{ij}^{(t)}
\left(s_i^{(t)},s_j^{(t)}\right)
=
\begin{cases}
\displaystyle
R_{C_i}^{+,(t)}
\left(
    \alpha
    \left[
        \Delta Q_{C_iC_j}^{(t)}
    \right]^+
    -
    T_{C_i}^{(t)}
    -
    O_{C_i}^{(t)}
\right),
&
\left(s_i^{(t)},s_j^{(t)}\right)=(C,C),
\\[4mm]
\displaystyle
R_{C_i}^{+,(t)}
\left(
    -O_{C_i}^{(t)}
\right),
&
\left(s_i^{(t)},s_j^{(t)}\right)=(C,D),
\\[4mm]
\displaystyle
R_{D_i}^{-,(t)}
\left(
    \beta
    \left[
        \Delta Q_{C_i}^{(t)}
    \right]^+
    -
    T_{D_i}^{(t)}
    -
    O_{D_i}^{(t)}
\right),
&
\left(s_i^{(t)},s_j^{(t)}\right)=(D,C),
\\[4mm]
\displaystyle
R_{D_i}^{-,(t)},
&
\left(s_i^{(t)},s_j^{(t)}\right)=(D,D).
\end{cases}
\label{eq:reputation-pairwise-payoff}
\end{equation}
Equivalently, for the first three interactions,
\begin{align}
\widetilde{\pi}_{ij}^{(t)}(C,C)
&=
R_{C_i}^{+,(t)}
\pi_{ij}^{(t)}(C,C),
\\
\widetilde{\pi}_{ij}^{(t)}(C,D)
&=
R_{C_i}^{+,(t)}
\pi_{ij}^{(t)}(C,D),
\\
\widetilde{\pi}_{ij}^{(t)}(D,C)
&=
R_{D_i}^{-,(t)}
\pi_{ij}^{(t)}(D,C).
\end{align}
Since the base payoff of the \((D,D)\) interaction is zero, the proposed
matrix directly assigns the reputation-dependent term
\(R_{D_i}^{-,(t)}\) to this interaction.
The accumulated reputation-adjusted payoff is therefore
\begin{equation}
\widetilde{\Pi}_i^{(t)}
=
\sum_{j\in\mathcal{N}_i}
\widetilde{\pi}_{ij}^{(t)}
\left(
    s_i^{(t)},s_j^{(t)}
\right),
\label{eq:reputation-accumulated-payoff}
\end{equation}
and the corresponding evolutionary fitness is
\begin{equation}
\widetilde{f}_i^{(t)}
\triangleq
\widetilde{\Pi}_i^{(t)}.
\label{eq:reputation-fitness}
\end{equation}

\subsection{Evolutionary Game Theory-Based Training Workflow} \label{workflow}
To strike a balance between individual model performance improvement and dynamic strategic adaptation, the proposed ÈFL framework organizes its runtime execution into discrete iterative rounds. The system operates on a $5 \times 5$ grid population where each spatial node acts as both a local machine learning agent and a strategic player in an evolutionary game. 

\begin{figure}[!htbp]
    \centering
    \includegraphics[width=0.75\linewidth]{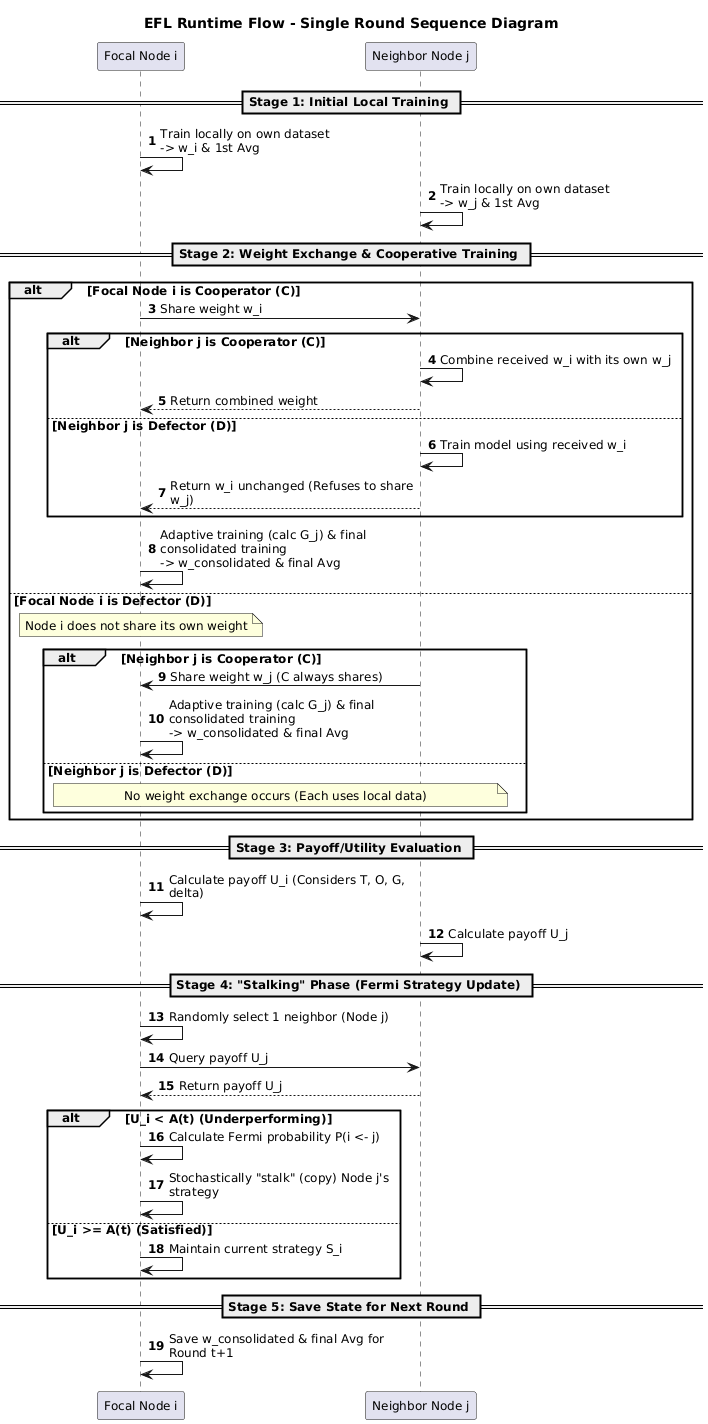}
    \caption{End-to-end workflow of the proposed EGT-enabled DFL architecture.}
    \label{fig:sequence_diagram}
\end{figure}

Depending on its current strategy - either Cooperator ($C$) or Defector ($D$), a node determines whether to contribute its parameter updates to its immediate spatial neighborhood. The complete operational workflow of a single execution round is detailed below and illustrated in Fig.~\ref{fig:sequence_diagram}.

\subsubsection{Workflow Breakdown}
\paragraph{Step 1: Initialization}
At generation $g = 1$, the Simulation Coordinator sets up the network $\mathcal{N}$, strategy space $\mathcal{S} \in \{C, D\}$, payoff matrix $\mathbf{M}$, Fermi intensity $\beta$, max generations $G$, and max rounds $R$. Each focal node $i$ is assigned an initial strategy $S_i \sim \text{Uniform}(\{C, D\})$ and a set of 4 spatial neighbors $\mathcal{N}_i$.

\paragraph{Step 2: Generation Loop Execution}
The evolutionary process runs for generation $g = 1 \dots G$ or \textbf{until DFL converges}. At the beginning of each generation $g$, cumulative payoffs are reset ($\Pi_i \leftarrow 0, \forall i \in \mathcal{N}$). Each generation consists of $R$ training rounds ($r = 1 \dots R$):

\paragraph{Step 3: Neighborhood Exchange Phase (Round $r$)}
For round $r = 1$, node $i$ performs initial self-training to obtain weight $w_i$. For subsequent rounds ($r > 1$), nodes bypass self-training and directly load the consolidated weights $w_{i,\text{consolidated}}$ from round $r-1$. The parameter exchange strictly adheres to node strategies $(S_i, S_j)$:
\begin{itemize}
    \item \textbf{Focal Node Strategy:} If focal node $i$ is a Cooperator ($C$), it transmits its current weight $w_i$ to its immediate neighbor node $j$. Conversely, if node $i$ is a Defector ($D$), it suppresses its parameters by sending returning unupdated weights.
    \item \textbf{Neighbor Response Strategy:} If neighbor node $j$ is a Cooperator ($C$), it combines the received weight $w_i$ with its own model parameters and returns the combined weight. If neighbor node $j$ is a Defector ($D$), it refuses parameter sharing and returns the received weight unchanged.
\end{itemize}

\paragraph{Step 4: Local Adaptive Training and Weight Consolidation}
Upon receiving returned weights from its spatial neighborhood $\mathcal{N}_i$, focal node $i$ executes local adaptive training to derive up to six performance evaluation averages:
\begin{itemize}
    \item \textit{Self-Training Average:} Obtained from the initial self-training pass (or prior state).
    \item \textit{Neighbor Evaluation Averages (up to 4):} Node $i$ trains separately on each weight returned from its 4 neighbors to calculate neighbor-specific performance averages and performance differentials $\Delta\text{Average}(G_{ji})$.
    \item \textit{Consolidated Average (1):} All received weights are aggregated together into a unified representation. The Local ML Trainer executes a final training pass on this aggregated model, returning the updated consolidated weight $w_{i,\text{consolidated}}$ and final consolidated performance metric.
\end{itemize}

\paragraph{Step 5: Game-Theoretic Payoff Accumulation}
Following model consolidation in round $r$, focal node $i$ calculates its round payoff $U_i^{(r)} = \sum_{j \in \mathcal{N}_i} \mathbf{M}(S_i, S_j)$ and updates its cumulative payoff:
\begin{equation}
    \Pi_i \leftarrow \Pi_i + U_i^{(r)}
\end{equation}

\paragraph{Step 6: Fermi Strategy Update}
After completing $R$ rounds in generation $g$, each node $i$ computes its average fitness $f_i = \frac{\Pi_i}{R_{\text{actual}}}$. Node $i$ then updates its strategy stochastically for the next generation:
\begin{itemize}
    \item Node $i$ uniformly samples one neighbor $j \sim \text{Uniform}(\mathcal{N}_i)$.
    \item Node $i$ computes the Fermi probability $P(S_i \leftarrow S_j)$ to decide whether to adopt neighbor $j$'s strategy:
    \begin{equation}
    P(S_i \leftarrow S_j) = \frac{1}{1 + e^{-\beta \cdot (f_j - f_i)}}
    \label{eq:fermi}
\end{equation}
    where $\beta \ge 0$ represents the selection pressure.
    \item Node $i$ updates its strategy $S_i$ accordingly for the next generation $g+1$.
\end{itemize}

\paragraph{Step 7: State Persistence and Convergence Check}
Finally, node $i$ retains $w_{i,\text{consolidated}}$ as starting inputs for the next iteration. The workflow continues until DFL reaches convergence.
\newpage

\begin{breakablealgorithm}
\caption{The $i^{th}$ generation evolutionary strategy under EGT in DFL}
\label{alg:smart_contract_coordination_opt}
\begin{algorithmic}[1]

\Statex \textbf{Initialization:} Node set $\mathcal{N}$, strategy space $\mathcal{S} \in \{C, D\}$, payoff matrix $A$, payoff $U$, fitness $\Pi$, Fermi intensity $\beta$, generations $G$, max rounds $R$.

\Comment{\textit{Randomly assign strategies to all nodes in the initial network.}}

\For{each node $i \in \mathcal{N}$}
    \State Assign random strategy $S_i \sim \text{Uniform}(\{C, D\})$ and set neighborhood $\mathcal{N}_i$ ($|\mathcal{N}_i| = 4$)
    \State Calculate the pay-off value of node $i$ in interactions with $\mathcal{N}_i$ neighbors
\EndFor

\For{generation $g = 1$ \textbf{to} $G$ or until DFL converges}
    \State Reset accumulate fitness $\Pi_i \leftarrow 0, \quad \forall i \in \mathcal{N}$
    \\
    \Comment{\textit{Calculate node's payoff}}
    \For{round $r = 1$ \textbf{to} $R$}
        \For{each node $i \in \mathcal{N}$}
            
            \State Calculate round payoff $U_i^{(r)} \leftarrow \sum_{j \in \mathcal{N}_i} A(S_i, S_j)$
            \State Calculate accumulate fitness $\Pi_g \leftarrow \Pi_i + U_i^{(r)}$
        \EndFor
    \EndFor
    \\
    \Comment{\textit{Strategy evolution}}
    
    \For{each node $i \in \mathcal{N}$}
        \State Compute average fitness $f_i \leftarrow \frac{\Pi_g}{R_{\text{actual}}}$
        \State Choose randomly a neighbor $j \sim \text{Uniform}(\mathcal{N}_i)$ (according to social event)
        \State Update strategy $S_i$ through Fermi probability to decide whether the strategy is updated in next generation:
        \Statex \makebox[\linewidth][c]{$P(S_i \leftarrow S_j) = \dfrac{1}{1 + e^{-\beta \cdot (f_j - f_i)}}$}
    \EndFor
\EndFor

\end{algorithmic}
\end{breakablealgorithm}

\begin{figure}[H]
    \centering
    \includegraphics[width=1\linewidth]{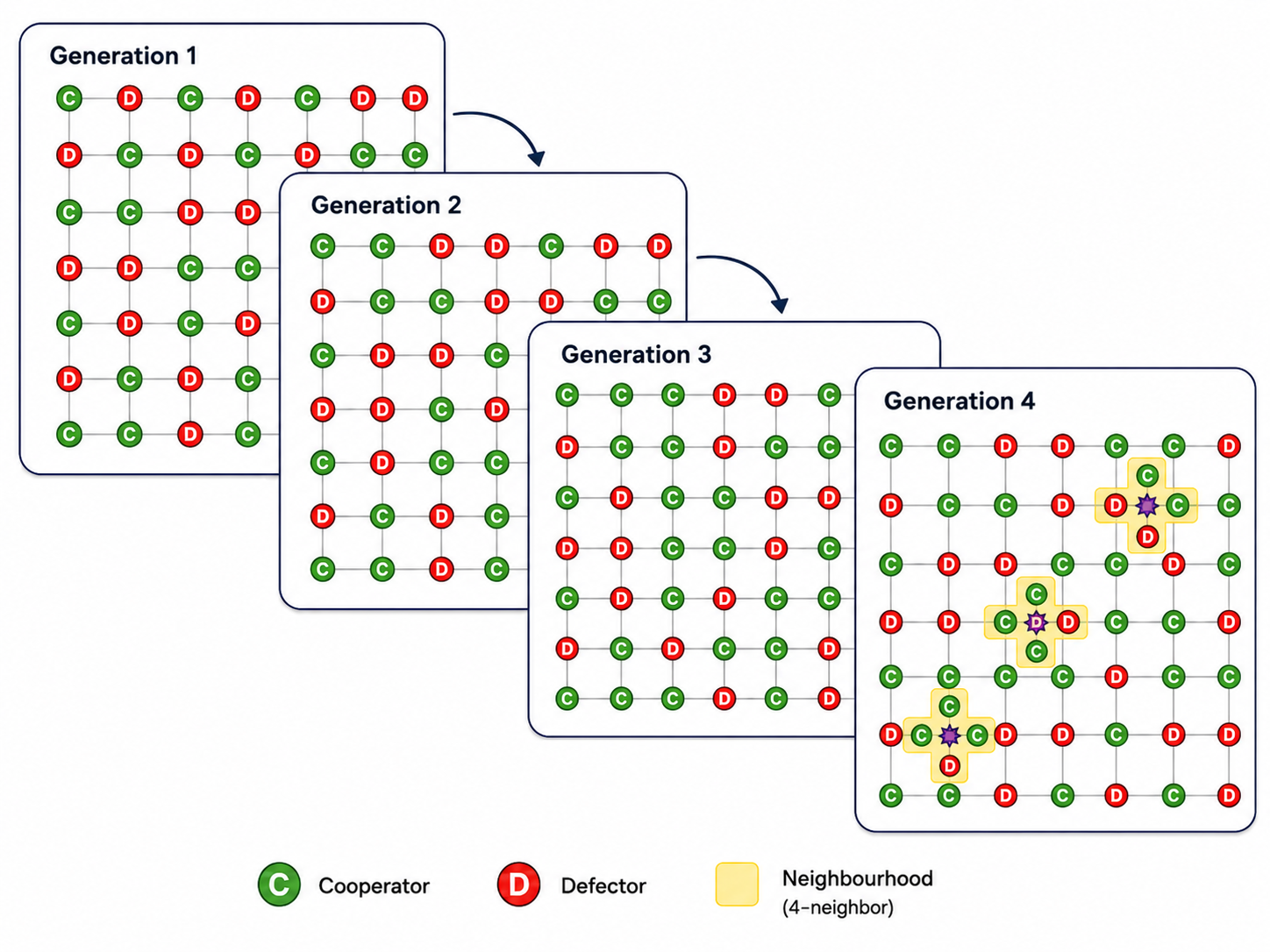}
    \caption{Evolution of node strategies across successive generations under the Fermi imitation mechanism. At each generation, a focal node compares its accumulated payoff with that of a randomly selected neighboring node and probabilistically adopts the neighbor's strategy according to their payoff difference, thereby driving the local transition between cooperation and defection over time.}
    \label{fig:fermi_imitation}
\end{figure}
\section{Experiments and Analysis}\label{sec5}

\subsection{Simulation Settings}\label{simulation}
To evaluate the proposed lattice Evolutionary-Game-Theoretic DFL framework, we construct a simulation environment in which each node acts both as a local learning agent and an evolutionary-game player. The experiment evaluates the evolution of node strategies and its effect on model performance.

The DFL network consists of a $n \times n$ periodic square lattice containing $n^2$ nodes. Each node interacts exclusively with its four immediate neighbors. At each round, a Cooperative ($C$) node contributes its model parameters to neighboring interactions, whereas a Defective ($D$) node limits its contribution while still receiving information from its neighbors.

After the neighborhood exchange and local training processes, each node calculates its cumulative payoff according to the payoff matrix defined in Section~\ref{architecture}. The node then randomly selects one neighboring node and applies the Fermi imitation rule to determine whether its strategy should be updated.

The main simulation parameters are summarized in Table~\ref{tab:sim_settings}.

\begin{table}[ht]
\centering
\caption{Common simulation settings}
\label{tab:sim_settings}
\renewcommand{\arraystretch}{1.15}
\begin{tabular}{ll}
\toprule
\textbf{Parameter} & \textbf{Setting} \\
\midrule
Network topology & $n\times n$ periodic square lattice \\
Lattice size & $n=50$ \\
Neighborhood size & 4 \\
Initial strategy distribution & Uniform random ($P(C)=P(D)=0.5$) \\
Maximum simulation rounds & 50 \\
Fermi selection parameter $\beta$ & 0.3 \\
Training cost $T$ & 3 \\
Communication cost $O$ & 0.5 \\
Synergy coefficient $\delta$ & $\delta_{max}=1.5$, $\delta_{min}=0.5$ \\
Reputation reward/penalty & $r=0.1$ \\
Evaluation metric & Accuracy  \\
\bottomrule
\end{tabular}
\end{table}

Three primary measurements are recorded during the simulation: the \textit{cooperation ratio}, overall model accuracy, and node fitness. The cooperation ratio is defined as
\begin{equation}
\rho_C(t)=\frac{N_C(t)}{N},
\end{equation}
where $N_C(t)$ is the number of cooperative nodes at round $t$ and $N=n^2$ is the total number of nodes. These measurements allow the experiment to evaluate both the evolutionary behavior of the network and its resulting learning performance.

Second, the model performance of the DFL system is monitored throughout the training process. The accuracy of the consolidated model is recorded after each round to determine whether the evolutionary dynamics result in improved learning performance and whether the model eventually approaches a stable state.

Third, the cumulative payoff of each node is recorded as its evolutionary fitness. This metric is used to examine whether cooperative or defective behavior is favored under the proposed payoff mechanism and to establish a relationship between strategic evolution and learning performance.

\subsection{Results and Evaluation}\label{result}

\begin{figure}[H]
    \centering
    \vspace{0.1cm} 
    \noindent \hspace{-9.3cm} \textbf{\textit{Baseline Scenario}}
    \nopagebreak
    \vspace{0.2cm}
    
    \begin{subfigure}[b]{0.23\textwidth}
        \centering
        \includegraphics[width=\textwidth]{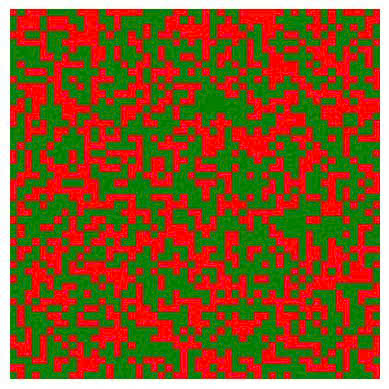}
        \label{fig:step_0}
    \end{subfigure}
    \hfill
    \begin{subfigure}[b]{0.23\textwidth}
        \centering
        \includegraphics[width=\textwidth]{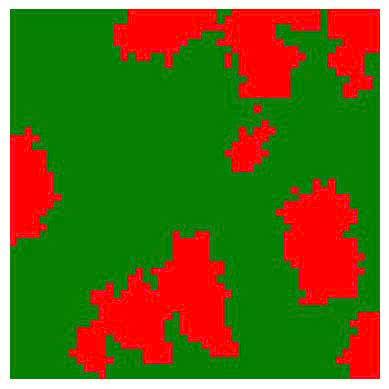}
        \label{fig:step_50}
    \end{subfigure}
    \hfill
    \begin{subfigure}[b]{0.23\textwidth}
        \centering
        \includegraphics[width=\textwidth]{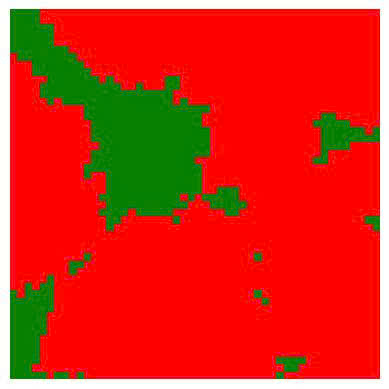}
        \label{fig:step_100}
    \end{subfigure}
    \hfill
    \begin{subfigure}[b]{0.23\textwidth}
        \centering
        \includegraphics[width=\textwidth]{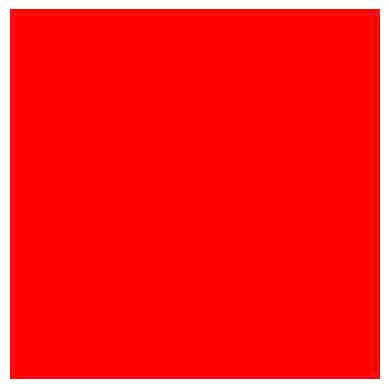}
        \label{fig:step_150}
    \end{subfigure}

    \vspace{0.1cm} 

    \noindent \hspace{-7.9cm} \textbf{\textit{Reputation-based Scenario}}
    \nopagebreak
    \vspace{0.2cm}

    \begin{subfigure}[b]{0.23\textwidth}
        \centering
        \includegraphics[width=\textwidth]{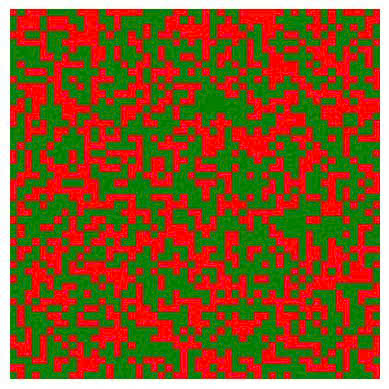}
        \label{fig:rep_step_0}
    \end{subfigure}
    \hfill
    \begin{subfigure}[b]{0.23\textwidth}
        \centering
        \includegraphics[width=\textwidth]{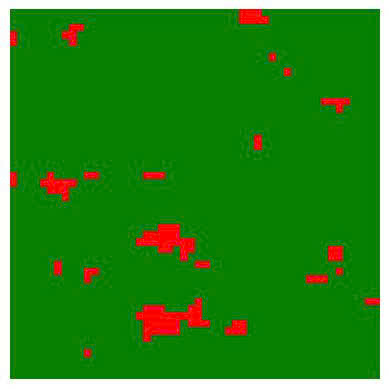}
        \label{fig:rep_step_50}
    \end{subfigure}
    \hfill
    \begin{subfigure}[b]{0.23\textwidth}
        \centering
        \includegraphics[width=\textwidth]{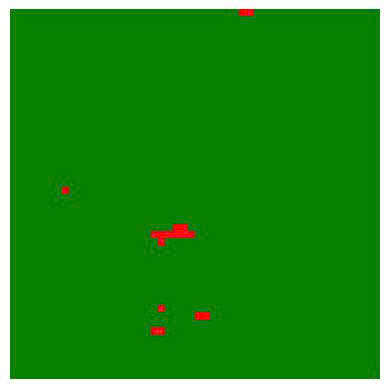}
        \label{fig:rep_step_100}
    \end{subfigure}
    \hfill
    \begin{subfigure}[b]{0.23\textwidth}
        \centering
        \includegraphics[width=\textwidth]{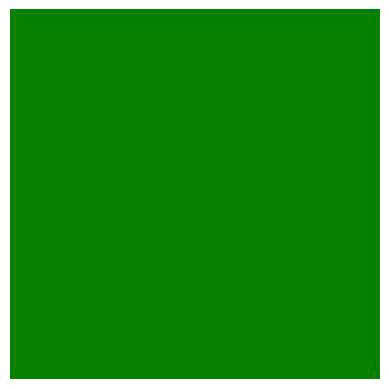}
        \label{fig:rep_step_150}
    \end{subfigure}

    \caption{Evolutionary process of strategy distribution with reputation and no reputation mechanism.}
    \label{fig:four_rounds_evolution}
\end{figure}

Figures~\ref{fig:four_rounds_evolution} illustrate the evolutionary dynamics of strategy adoption under the Fermi imitation process. Each snapshot represents the distribution of strategies at different generations, where red denotes defectors and green denotes cooperators.

\textbf{Without the reputation mechanism}, defectors gradually dominate the population. This behavior arises from the proposed payoff formulation, in which the reward is scaled by the model's improvement in accuracy $\left(\Delta Q\right)$. During the early stages of training, cooperation produces substantial accuracy gains, making collaborative training beneficial. However, as the models approach convergence, the marginal improvement in accuracy diminishes. Consequently, the benefit obtained from additional cooperation becomes insufficient to justify for its computational and communication costs. Defection therefore creates a higher net payoff, and under the Fermi imitation rule, neighboring players increasingly adopt the defection strategy, eventually leading the population to a fully defective state.

\textbf{With the reputation mechanism}, the evolutionary outcome is reversed, with cooperation gradually dominating the population. Although the marginal improvement in model accuracy $\left(\Delta Q\right)$ decreases as training progresses, the proposed payoff matrix incorporates each player's reputation score as an additional incentive. Cooperative players continuously accumulate reputation through repeated participation, resulting in a progressively larger payoff bonus. As training converges, the growth of the reputation reward surpasses the diminishing contribution of $\Delta Q$ to the payoff. Therefore, cooperation remains the more profitable strategy despite the reduced marginal learning benefit. Under the Fermi imitation rule, neighboring agents are increasingly likely to adopt the cooperative strategy, leading to the expansion of cooperative clusters until the population converges to an almost entirely cooperative state.

\begin{figure}[H]
    \centering
    \begin{subfigure}[b]{0.45\textwidth}
        \centering
        \includegraphics[width=\textwidth]{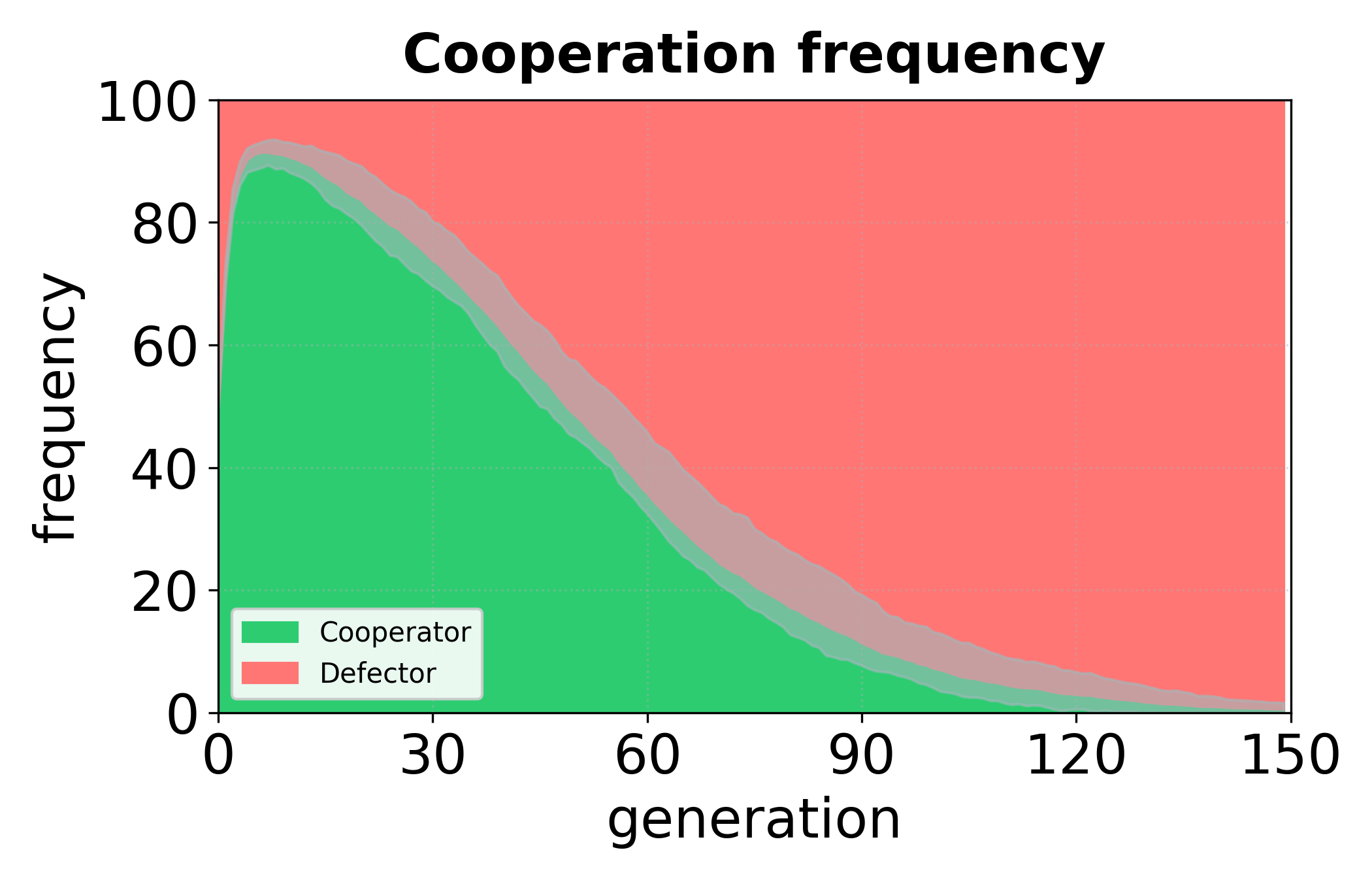}
     
        \label{fig:img1}
    \end{subfigure}
    \hfill
    \begin{subfigure}[b]{0.45\textwidth}
        \centering
        \includegraphics[width=\textwidth]{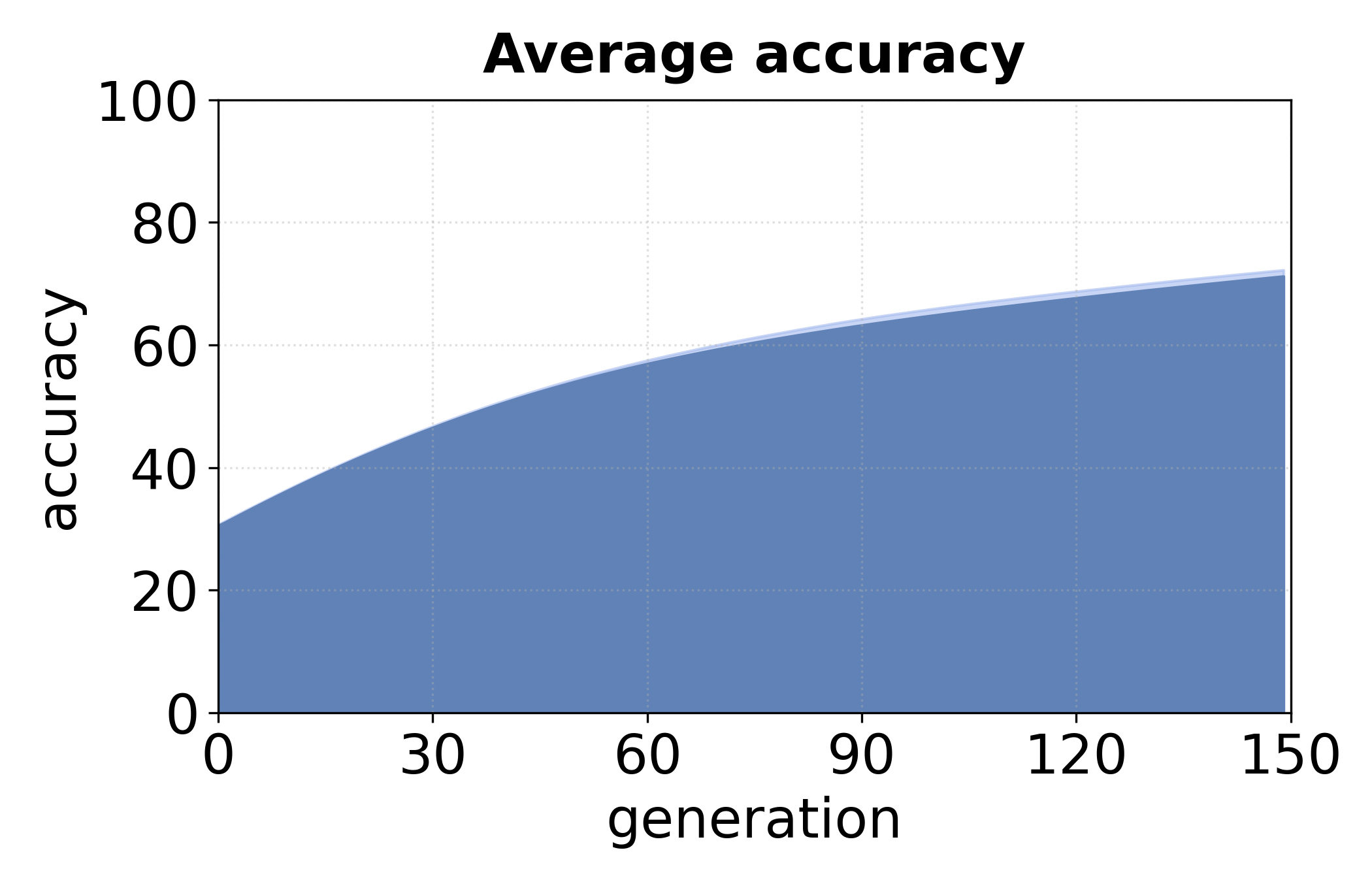}
       
        \label{fig:img2}
    \end{subfigure}

    \begin{subfigure}[b]{0.45\textwidth}
        \centering
        \includegraphics[width=\textwidth]{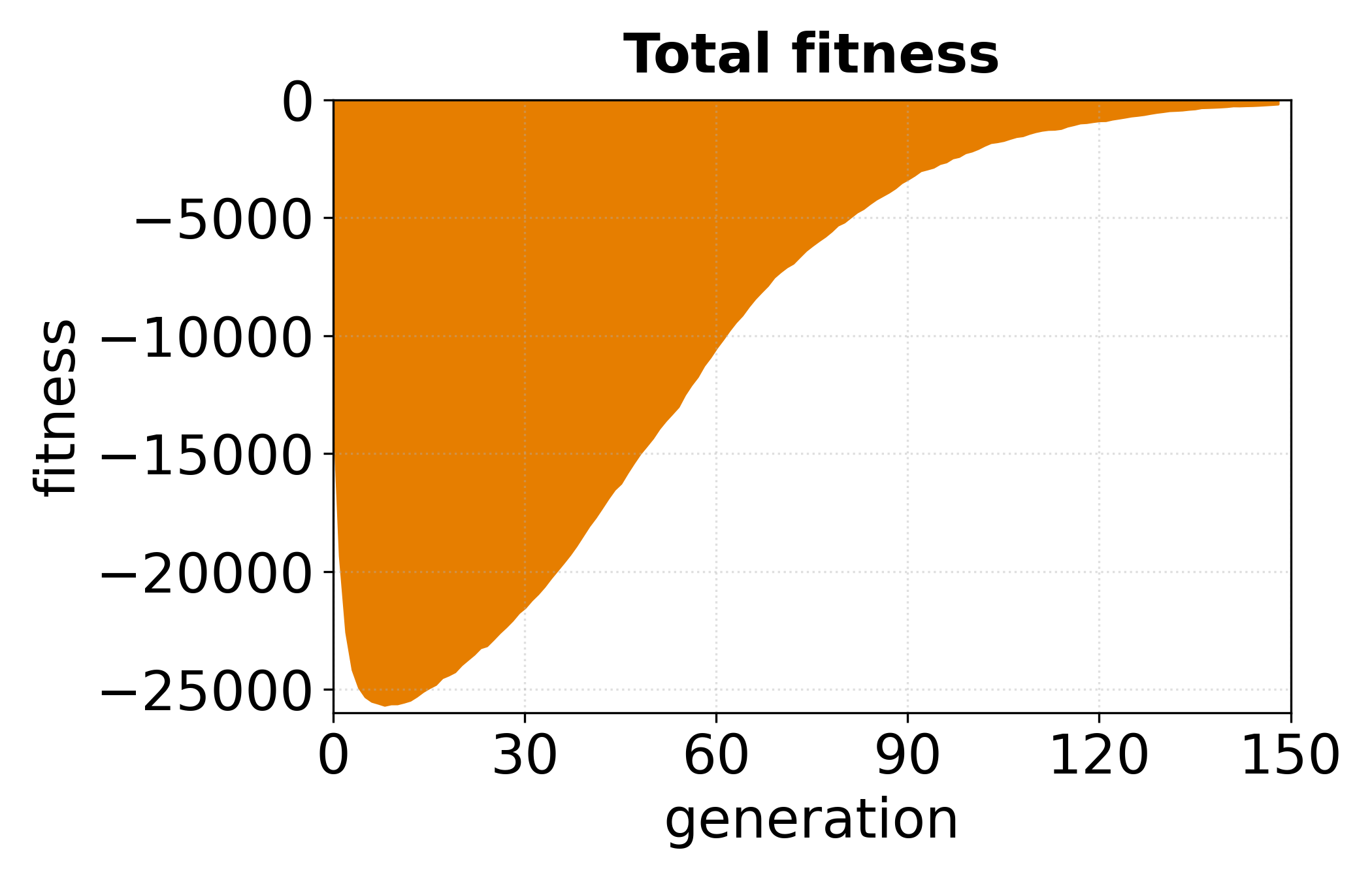}

        \label{fig:img3}
    \end{subfigure}
    \hfill
    \begin{subfigure}[b]{0.45\textwidth}
        \centering
        \includegraphics[width=\textwidth]{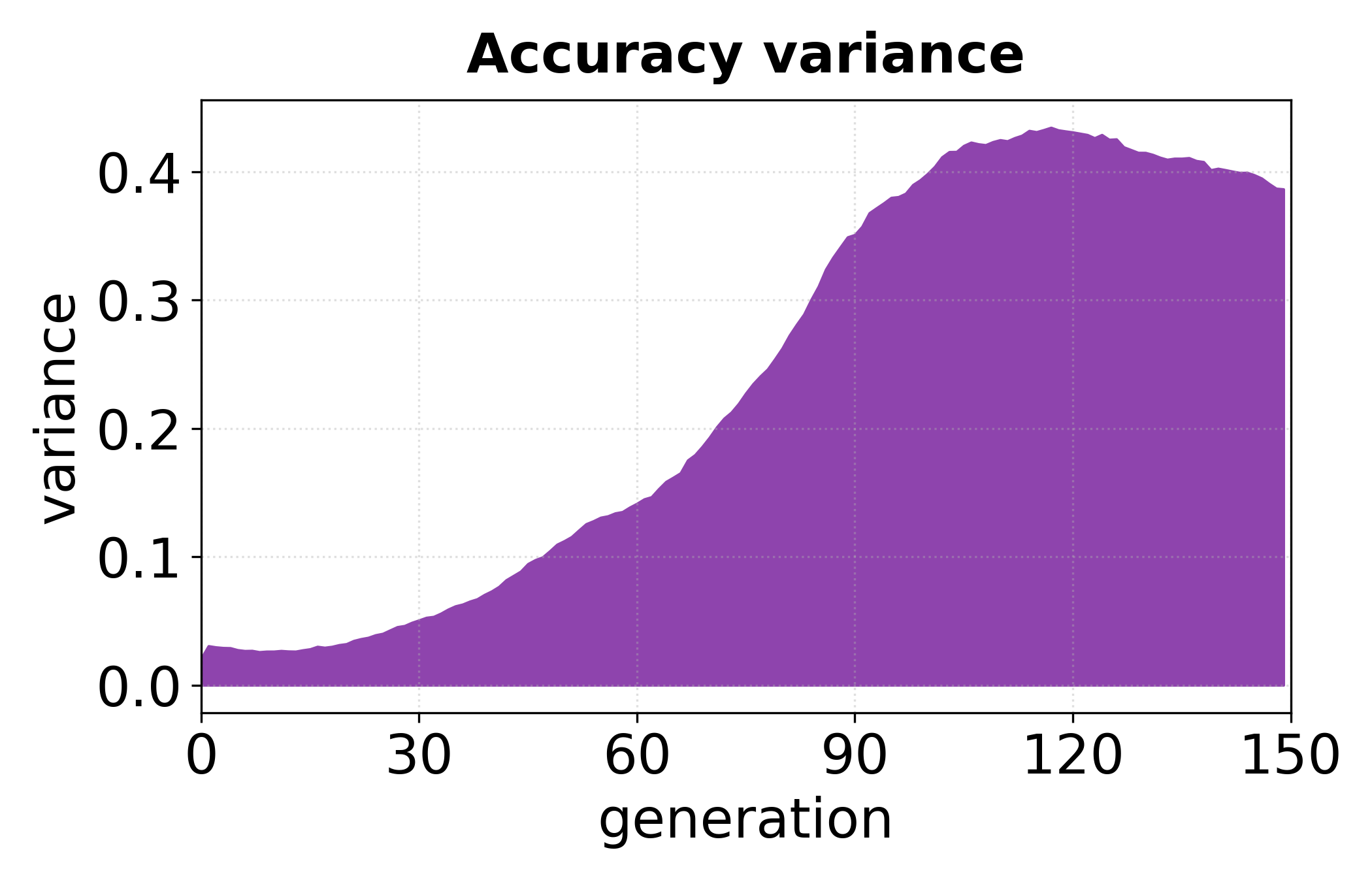}

        \label{fig:img4}
    \end{subfigure}
    \caption{Comparison of strategy distribution over time with no reputation mechanism.}
    \label{fig:six_images_2x2}
\end{figure}

Figure \ref{fig:six_images_2x2} illustrates the evolutionary dynamics \textbf{without the reputation mechanism}. Initially, cooperation is beneficial because collaborative training produces substantial improvements in model accuracy $\left(\Delta Q\right)$. However, as the global model approaches convergence, the marginal accuracy gain gradually diminishes, making the communication and computation costs of cooperation outweigh its benefits. Consequently, defection becomes the more profitable strategy, and the Fermi imitation process drives the population toward an almost entirely defective state.

The decline in cooperation is reflected in the remaining performance metrics. As fewer players participate in training, the overall training cost decreases, resulting in a \textit{gradual recovery} of the total fitness. However, reduced collaboration also limits knowledge sharing, leading to a lower final average accuracy. Meanwhile, the increasing accuracy variance indicates that individual models become \textit{more heterogeneous}, as cooperative players continue improving while defective players largely cease updating their models.

\begin{figure}[H]
\centering

    \begin{subfigure}[b]{0.45\textwidth}
        \centering
        \includegraphics[width=\textwidth]{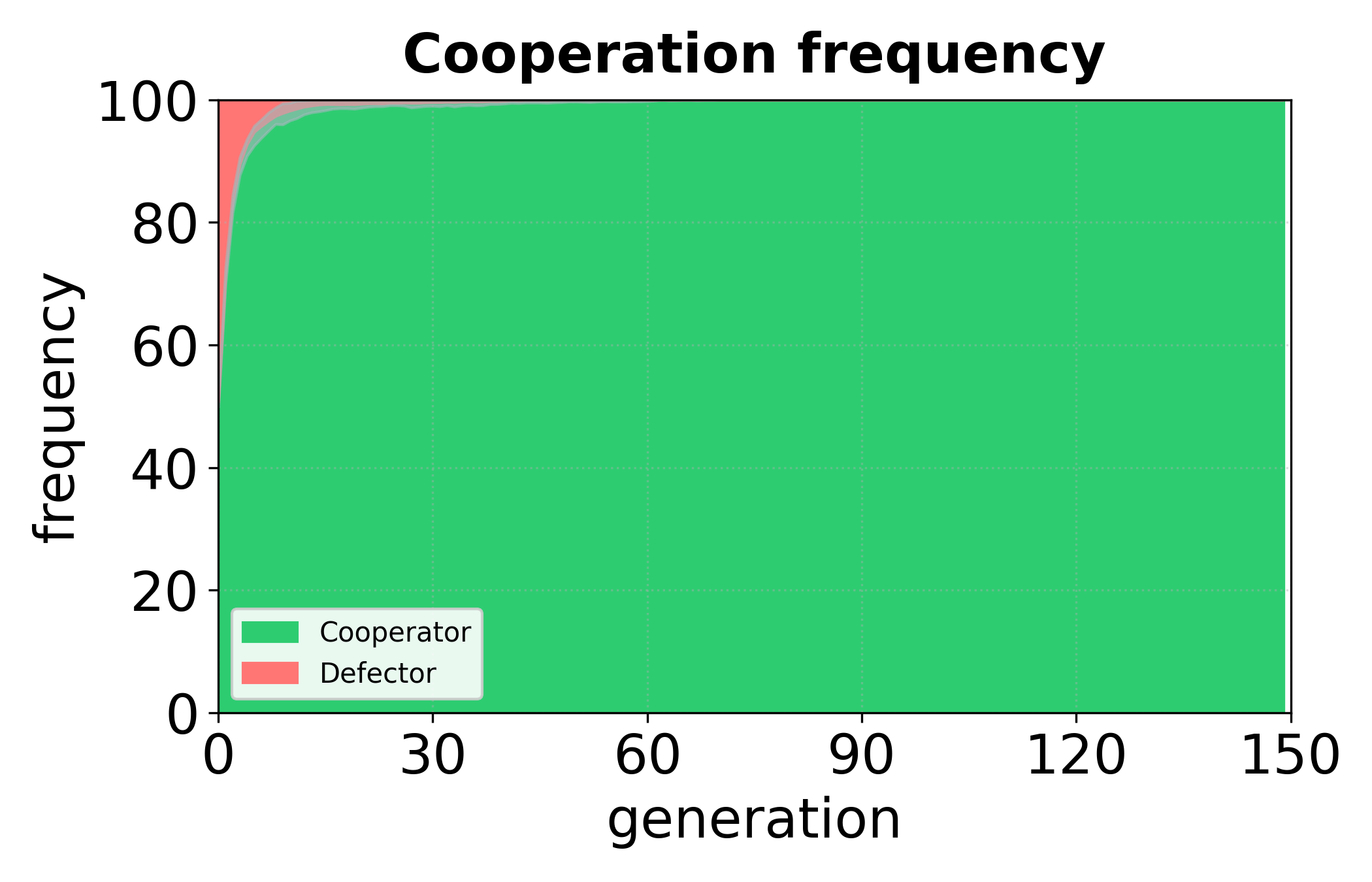}
      
        \label{fig:img5}
    \end{subfigure}
    \hfill
    \begin{subfigure}[b]{0.45\textwidth}
        \centering
        \includegraphics[width=\textwidth]{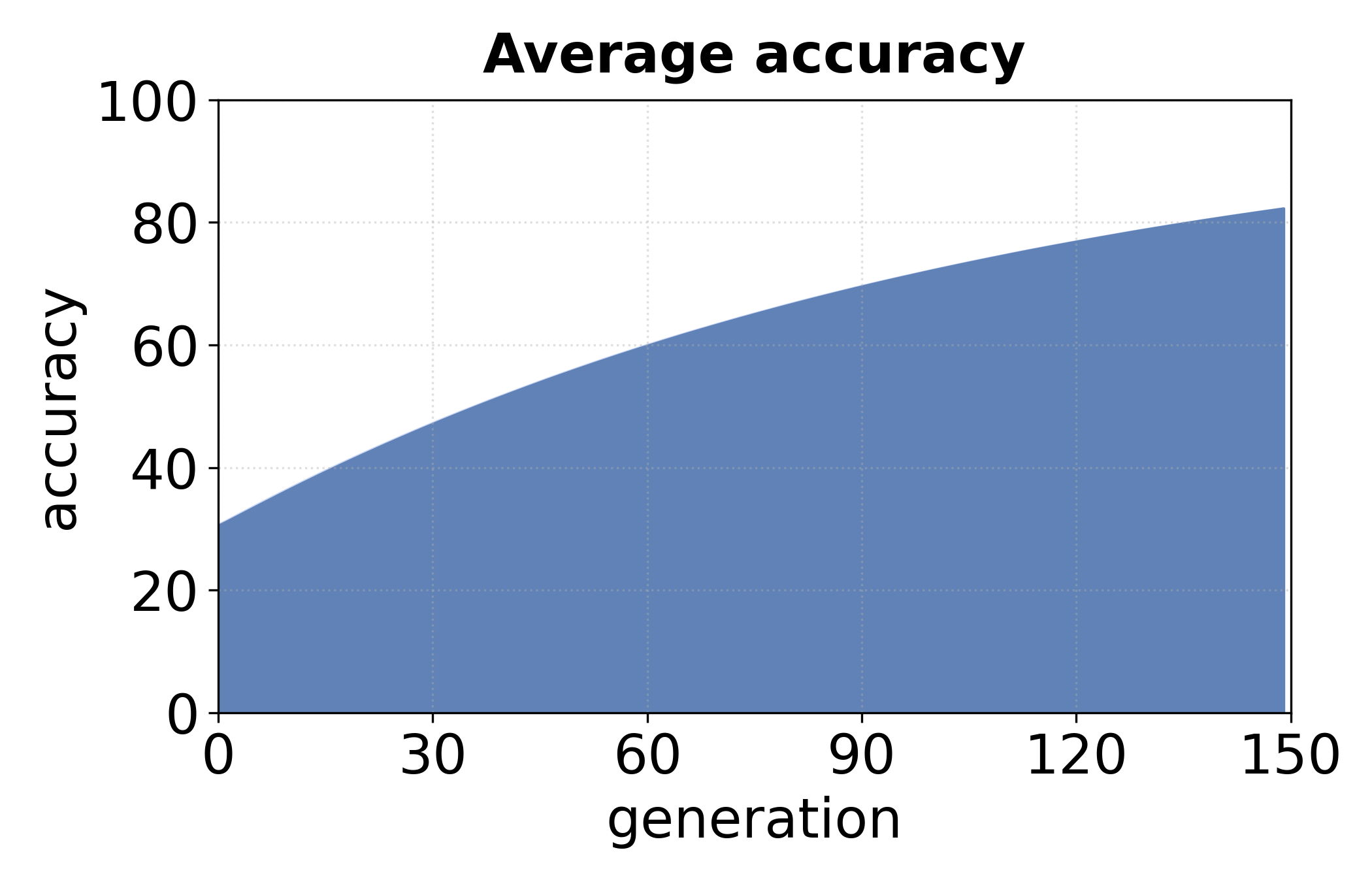}
        
        \label{fig:img6}
    \end{subfigure}
    \hfill
    \begin{subfigure}[b]{0.45\textwidth}
        \centering
        \includegraphics[width=\textwidth]{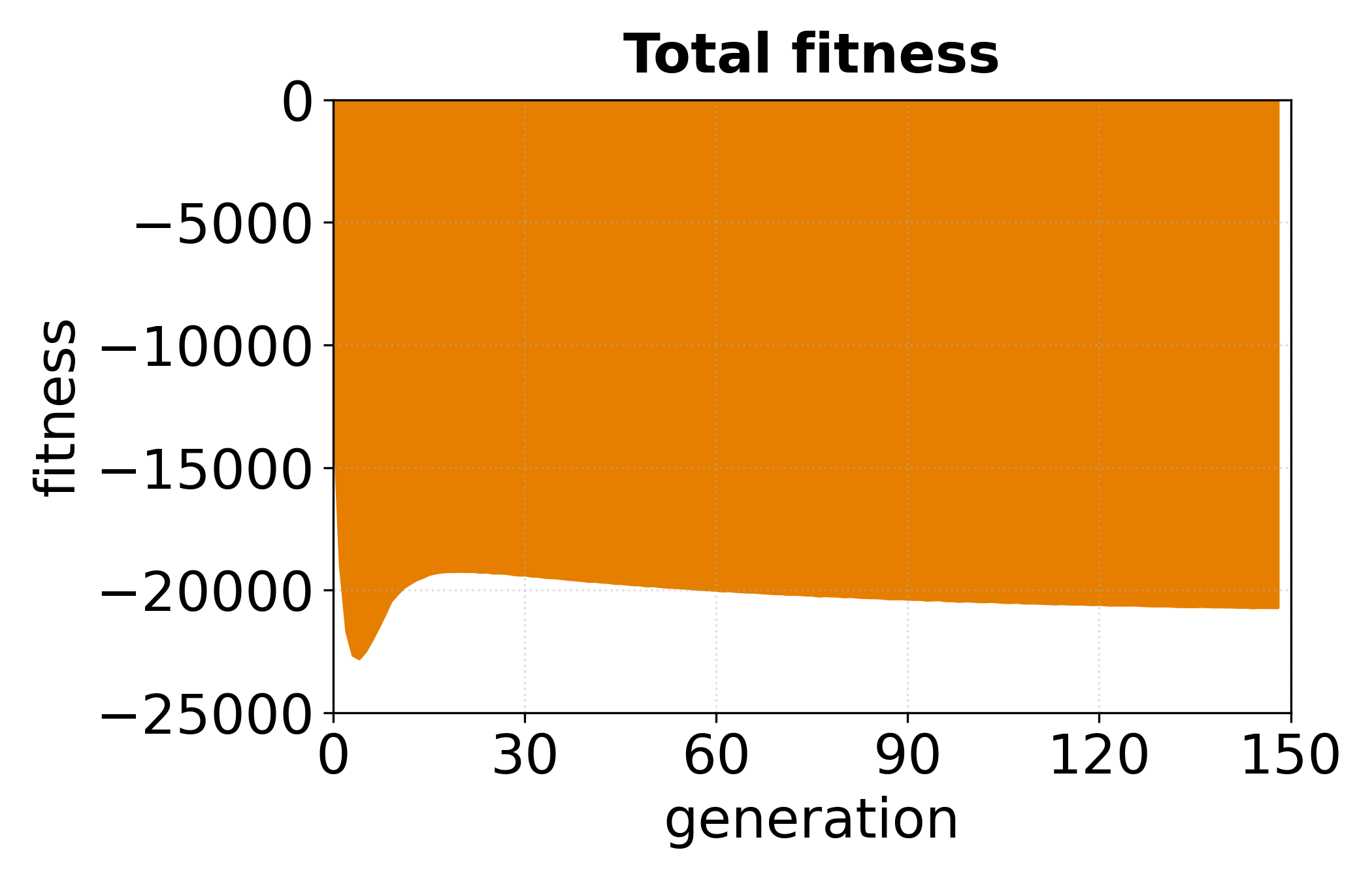}
       
        \label{fig:img7}
    \end{subfigure}
    \hfill
    \begin{subfigure}[b]{0.45\textwidth}
        \centering
        \includegraphics[width=\textwidth]{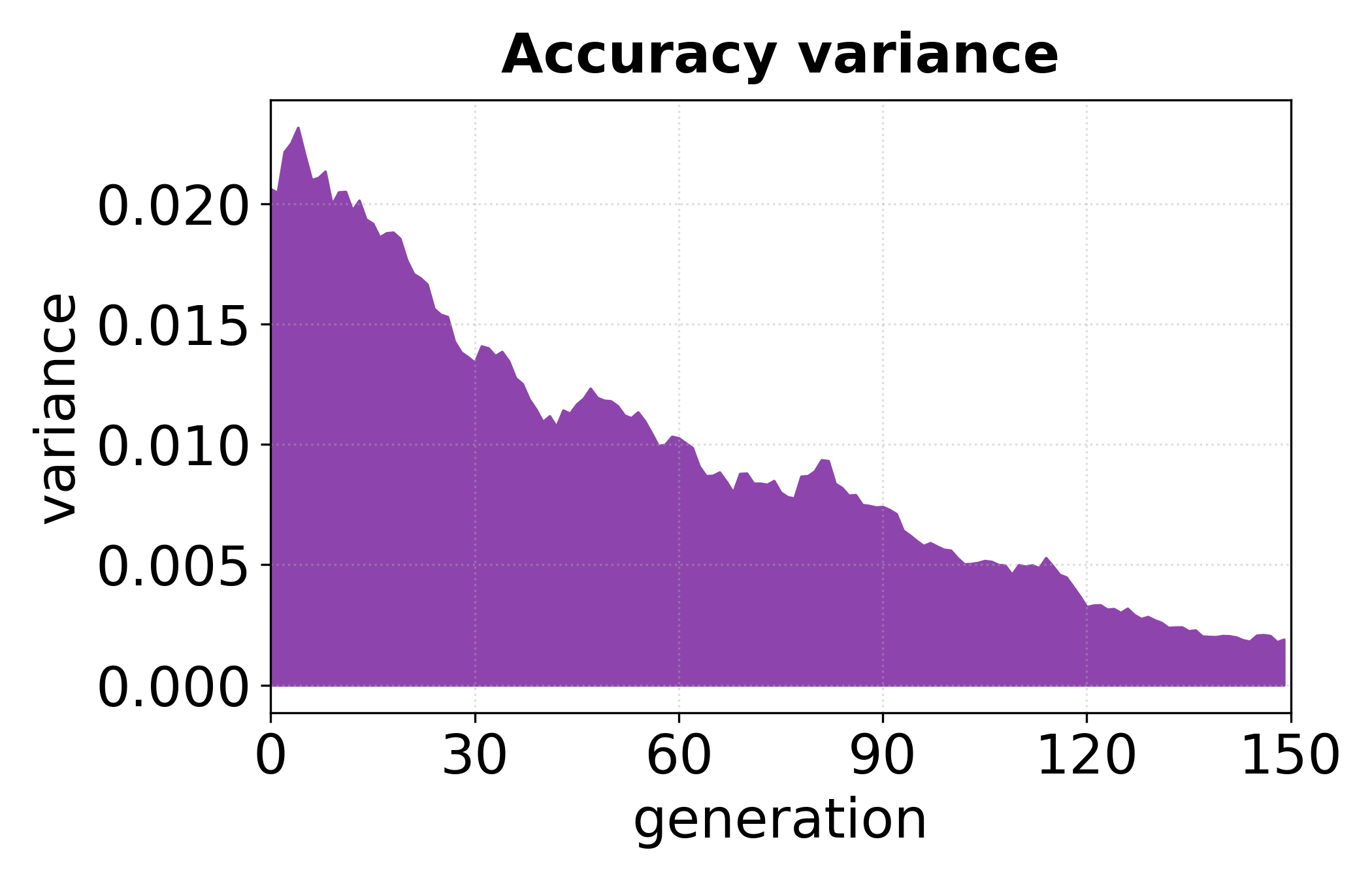}
       
        \label{fig:img8}
    \end{subfigure}
    \caption{Comparison of strategy distribution over time with reputation mechanism.}
    \label{fig:six_images_3x2}
\end{figure}

Figure \ref{fig:six_images_3x2} presents the evolutionary dynamics \textbf{when the reputation mechanism is incorporated} into the payoff formulation. Although the improvement in model accuracy $\left(\Delta Q\right)$ gradually diminishes as training progresses, cooperative players continuously accumulate reputation through repeated participation. As the reputation reward eventually outweighs the diminishing contribution of $\left(\Delta Q\right)$, cooperation remains the more profitable strategy. Consequently, the Fermi imitation process rapidly drives the population toward an almost entirely cooperative state.

Sustained cooperation enables continuous knowledge sharing among participants, resulting in a \textit{higher final} average model accuracy than the baseline. Although maintaining cooperation incurs additional communication and computation costs, the total fitness remains relatively stable due to the reputation incentive. Furthermore, the accuracy variance \textit{steadily decreases} throughout the evolutionary process, indicating that the distributed models progressively converge to similar performance levels. These results demonstrate that the reputation mechanism not only improves learning performance but also promotes a more consistent and robust decentralized learning process.

\section{Conclusion and Discussion}\label{sec6}

In this study, we model the Decentralized Federated Learning network under a lattice structure, where each node represents a participating agent seeking to maximize its own utility. By establishing a bounded-rationality agent model with spatial-cluster interactions, this research enables the intuitive observation of strategic evolution and the behavioral propagation dynamics across the system. Grounded in Evolutionary Game Theory (EGT), we analyze the strategic transition dynamics and formulate a comprehensive payoff matrix. Furthermore, to foster sustainable cooperation and deter free-riding behaviors, a reputation-based reward-and-punishment mechanism is integrated into the framework. Experimental results not only clearly illustrate the spatial propagation patterns of behavioral interactions, but also validate the superior effectiveness of the reputation-based reward-and-punishment mechanism in optimizing overall performance and suppressing the free-riding problem in DFL.

In future work, we intend to extend this framework to more complex network topologies, such as scale-free and small-world networks, to assess the impact of network connectivity on system behavior. Additionally, future extensions will incorporate realistic client connectivity parameters (e.g., latency and bandwidth constraints) and explore multi-state models to accommodate the dynamic and flexible roles of individual nodes.

\begin{credits}
\subsubsection{\ackname} We acknowledge Ho Chi Minh City University of Technology (HCMUT), VNU-HCM for supporting this study.
 TAH is supported by EPSRC (grant EP/Y00857X/1). MHD was supported by EPSRC grant EP/Y008561/1. 
T.A.H. acknowledges travel and accommodation support from the Ho Chi Minh City University of Technology (HCMUT), VNUHCM (Adjunct Professorship scheme HCMUT\text{-}VNUHCM). L.H.T and T.A.T.N acknowledge the Ho Chi Minh City University of Technology (HCMUT), VNUHCM for supporting this study.

\end{credits}
%
%
%
\bibliographystyle{splncs04}
\bibliography{references}
%





\end{document}